%% file: AnonymousSubmission2027.tex
\documentclass[letterpaper]{article} % DO NOT CHANGE THIS
\usepackage[preprint]{aaai2027}  % DO NOT CHANGE THIS
\usepackage[hyphens]{url}  % DO NOT CHANGE THIS
\usepackage{graphicx} % DO NOT CHANGE THIS
\usepackage{natbib}  % DO NOT CHANGE THIS AND DO NOT ADD ANY OPTIONS TO IT
\usepackage{caption} % DO NOT CHANGE THIS AND DO NOT ADD ANY OPTIONS TO IT
\usepackage{algorithm}
\usepackage{algorithmic}

\usepackage{newfloat}
\usepackage{listings}
\DeclareCaptionStyle{ruled}{labelfont=normalfont,labelsep=colon,strut=off} % DO NOT CHANGE THIS
\floatstyle{ruled}
\newfloat{listing}{tb}{lst}{}
\floatname{listing}{Listing}

\usepackage{booktabs}

\usepackage{adjustbox}

\usepackage{tabularx}
\usepackage{multirow}
\usepackage{makecell}
\usepackage{fancyvrb,fvextra}

\usepackage{subfig}

\usepackage{xurl}

\usepackage{dsfont}
\usepackage{amsfonts}
\usepackage{amsmath}
\usepackage{mathtools}
\usepackage{stackrel}
\DeclareMathOperator*{\argmax}{arg\,max}

\usepackage{listings}
\usepackage{here}

\usepackage[english]{babel}

\usepackage{xcolor}
\usepackage{mdframed}
\newmdenv[
  linecolor=gray!55,
  linewidth=0.5pt,
  backgroundcolor=white,
  innerleftmargin=3mm,
  innerrightmargin=3mm,
  innertopmargin=2.5mm,
  innerbottommargin=2.5mm,
  skipabove=8pt,
  skipbelow=8pt,
]{examplebox}

\title{TalkMatrix: Structured Multi-Prompt Selection for Consistent and Diverse Character Dialogue}
\author{
    Ayuto Tsutsumi\textsuperscript{\rm 1, \rm 2},
    Yuu Jinnai\textsuperscript{\rm 2}
}
\affiliations{
    \textsuperscript{\rm 1}Tokyo Metropolitan University\\
    \textsuperscript{\rm 2}CyberAgent\\
    tsutsumi-ayuto@ed.tmu.ac.jp, ddyuudd@gmail.com
}

\begin{document}

\maketitle

\begin{abstract}
Candidate-based decoding typically selects a completion for each prompt independently, but many applications require a collection of outputs that satisfies global, non-decomposable requirements.
We formulate this setting as structured multi-prompt, multi-completion selection: given a candidate pool for every prompt, select one completion per prompt to optimize a collection-level objective.
We instantiate the problem in character dialogue, where each character should remain consistent across situations, each line should fit its situation, and characters and situations should remain distinguishable.
Our method, TalkMatrix, generates multiple candidates for every character--situation pair and jointly selects a complete matrix using four embedding-based consistency and diversity objectives.
Because a weighted sum can improve some dimensions by sacrificing another, TalkMatrix maximizes the worst-performing objective through a two-level minimax formulation.
We approximately optimize the resulting discrete objective with multi-start coordinate ascent, and compare it with local, partial-matrix, and generic combinatorial search baselines.
We run experiments on $50$ synthetic role-playing scenarios and $25$ curated board game scenarios where multiple characters interact in predefined situations.
An LLM-as-a-judge rates matrix-level selection higher than random and independent cell-level selection baselines.
These results show the value of structured selection for globally controlled dialogue generation, while our empirical validation remains specific to role-playing scenarios.
\end{abstract}

\section{Introduction}
Sampling from an LLM produces locally plausible candidates, and candidate-based decoding can select a strong completion for a single prompt~\cite{goel2000minimum,eikema-aziz-2020-map}. Many applications, however, require a collection of outputs whose quality depends on relations among the selected completions. If the collection-level objective contains consistency, diversity, coverage, or other cross-prompt terms, selecting the best candidate for each prompt independently may be insufficient. We study this non-decomposable setting as \emph{structured multi-prompt, multi-completion selection} where one completion is selected from each prompt's candidate pool to optimize the quality of the complete collection.

Narrative dialogue is a representative instance. A line must fit its local situation, all lines for one character should express a coherent persona, and different characters and situations should remain distinguishable. These requirements interact across the complete set of lines and therefore create a coupled combinatorial selection problem.

This problem arises in story-driven games and other entertainment media, where manually writing dialogue for many characters across many situations is costly. LLM-based role-playing agents offer a scalable alternative \cite{wang-etal-2024-rolellm,wang2025opencharacter,huang2025interactive,ruangtanusak2025talkless}, but generation alone does not ensure that the resulting collection is balanced across all of the requirements above. We focus on scripted narrative settings in which a fixed set of characters appears in multiple predefined situations, and cast dialogue construction as joint selection from LLM-generated candidates.

This paper presents {\bf TalkMatrix}, a structured selection method for scalable and globally controlled dialogue generation. Our contributions are as follows. (1) We formulate structured multi-prompt, multi-completion selection as a non-decomposable combinatorial optimization problem over candidate pools, and instantiate it as a character--situation matrix. (2) For this dialogue setting, we define four embedding-based consistency and diversity objectives and introduce a two-level minimax formulation that protects the worst-performing character, situation, and quality dimension instead of allowing strong dimensions to compensate for weak ones. (3) We develop a GPU-parallel, multi-start coordinate-ascent procedure that exploits the row--column structure of the objective and returns the best solution found, without claiming global optimality. (4) Through language-model-judge evaluation on $50$ role-playing scenarios and additional selection-policy baselines on the $25$ curated scenarios, we show that the main gain comes from optimizing a matrix-level objective over multiple cells, rather than from independent cell-level reranking. 
The formulation is general, whereas our objectives and empirical evidence concern character dialogue.

\begin{figure*}[t]
    \centering
    \includegraphics[width=\textwidth]{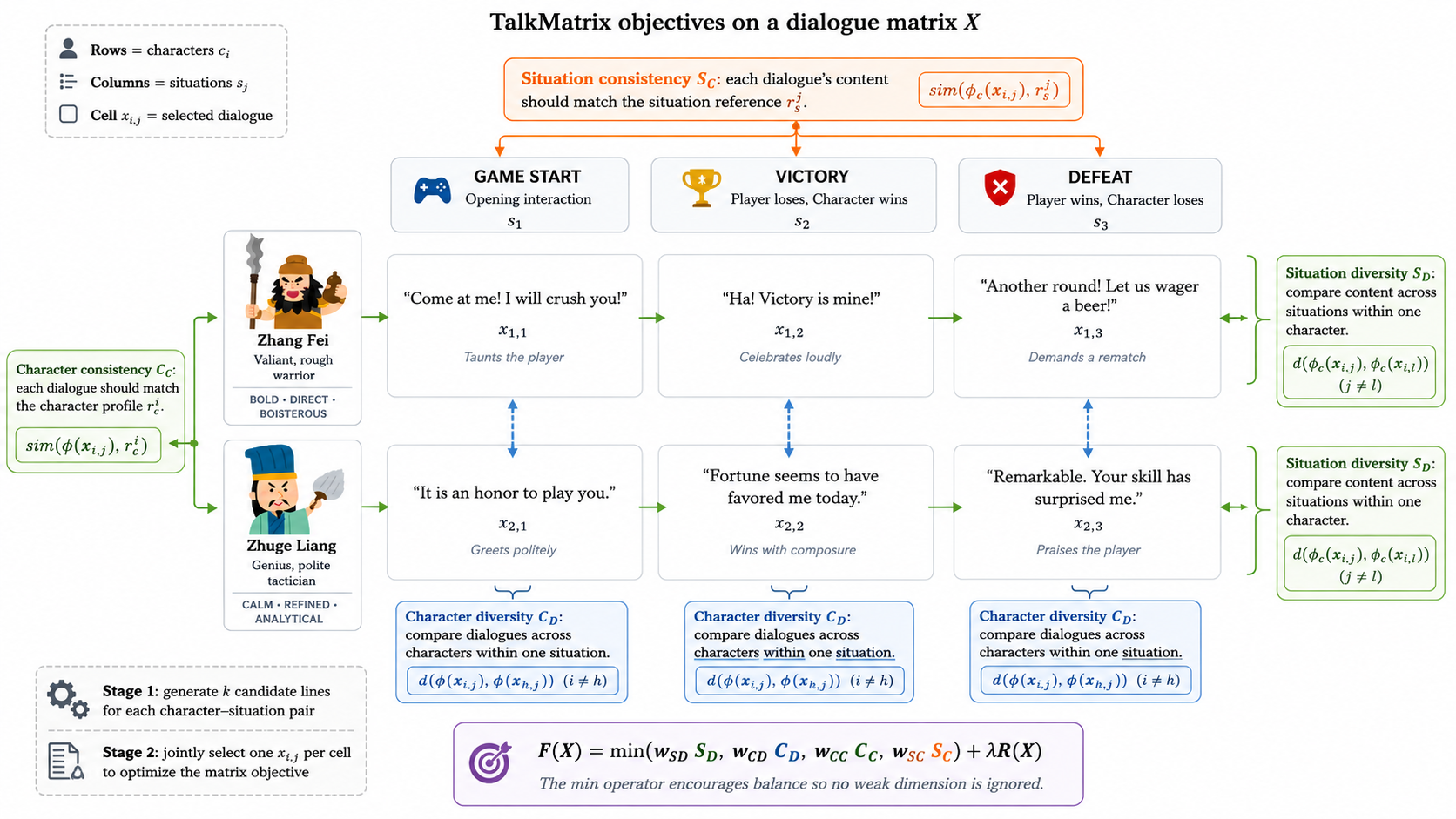}
    \caption{High-level goal of TalkMatrix. Each row represents one character across situations, and each column represents one situation across characters. TalkMatrix jointly selects dialogue lines that preserve each character's persona and tone (character consistency), fit the intended situation (situation consistency), distinguish characters in the same situation (character diversity), and vary appropriately across situations (situation diversity).}
    \label{fig:talkmatrix_diagram}
\end{figure*}

\section{Background}

\paragraph{Text generation problem.}
A standard text generation problem is to select a completion $x$ from a set of possible completions $\mathcal{Y}$ given a prompt $p$. A common approach to generate $x$ from a language model $P(x|p)$ is to sample randomly according to $P(x|p)$ or to use beam search to find a high-probability completion. The goal is to maximize a scoring function $f(x)$ that measures the quality of the completion:
\begin{equation}
x = \argmax_{x \in \mathcal{Y}} f(x).
\label{eq:selection}
\end{equation} 
These methods are local to one prompt and do not consider the quality of a collection of completions across multiple prompts. 
% For a prompt $p$, candidate-based decoding first constructs a finite set $\mathcal{Y}_p=\{y_p^1,\ldots,y_p^k\}$ and then selects one completion. 
% Minimum Bayes risk (MBR) decoding~\cite{goel2000minimum,eikema-aziz-2020-map} is a representative approach: under an empirical distribution over candidates and a utility function $u$, it selects
% \begin{equation}
% \hat x_p=\argmax_{x\in\mathcal{Y}_p}\frac{1}{k}\sum_{x'\in\mathcal{Y}_p}u(x,x').
% \label{eq:single-mbr}
% \end{equation}
% This decision is local to one prompt. 
Now consider a set of related prompts $p_1,\ldots,p_N$, with candidate set $\mathcal{Y}_q$ for each $p_q$. 
Structured multi-prompt selection chooses one candidate for every prompt,
\begin{equation}
\hat{\mathbf x}=\argmax_{\mathbf x\in\mathcal{Y}_1\times\cdots\times\mathcal{Y}_N}F(\mathbf x).
\label{eq:general-selection}
\end{equation}
When $F(\mathbf x)=\sum_q f_q(x_q)$, the problem decomposes into independent single-prompt decisions. We study the non-decomposable case, in which $F$ also contains cross-prompt terms and the quality of one selection depends on candidates chosen for other prompts. This formulation concerns post-hoc selection from independently generated pools; it does not require the language model to jointly generate all completions.

\paragraph{Persona-based dialogue generation.}
Early work conditions dialogue agents on explicit persona profiles to make responses more specific and consistent \cite{zhang-etal-2018-personalizing}, and later studies use character profiles or traits to condition language models and generate responses that are consistent with a given persona \cite{zheng2019personalized,chen-etal-2023-towards-robust}. Subsequent work augments personas with commonsense knowledge to sustain consistent and engaging narratives \cite{gao-etal-2023-peacok} and maintains persona over long-term interactions \cite{li-etal-2025-hello}. These models are mostly trained on open-domain datasets and aim to maintain personality traits over long dialogues.

\paragraph{Role-playing agents.}
LLM-based role-playing agents (RPAs) portray specified characters for applications ranging from entertainment to NPC dialogue in games \cite{wang-etal-2024-rolellm,tao-etal-2024-rolecraft,huang2025interactive,ruangtanusak2025talkless}. Recent efforts enhance RPAs' fidelity to a character's linguistic style \cite{chen2024multitask}, extend them to multi-character settings \cite{yu-etal-2024-neeko}, and support configurable or customizable characters \cite{he-etal-2025-crab,wang2025opencharacter}. Unlike these works, which condition generation on a single character or persona, we jointly select dialogue lines across an entire matrix of characters and situations so that the selected lines are simultaneously consistent and diverse.

\paragraph{Evaluating role-playing agents.}
A growing body of benchmarks evaluates RPAs along dimensions such as role knowledge, persona consistency, and conversational ability \cite{wang-etal-2024-rolellm,dmtrolebench2025,wu-etal-2025-raiden,he-etal-2025-crab,ngokpol2025worldbench}, emotional fidelity \cite{feng-etal-2025-emocharacter}, plot-progression capability \cite{zhang-etal-2025-roleplot}, and the gap between literary and real-world dialogue \cite{yin-etal-2025-charactercraft}. {\bf Following this line of work, we assess generated dialogue matrices with an LLM-as-a-judge protocol.}

\paragraph{Diversity in generated dialogue.}
Many works address the problem of generating diverse responses by modifying decoding strategies \cite{Vijayakumar2016,ippolito-etal-2019-comparison,Holtzman2020The,jinnai-etal-2024-generating,li2024dtllm} or by encouraging variation through training objectives \cite{chung-etal-2023-increasing,wang2024beyond,guo2024benchmarking,lanchantin2025diverse}. 
In contrast, our focus is on maximizing diversity across characters and situations within a fixed set of candidate lines.

\section{TalkMatrix: Structured Multi-Prompt Selection for Character Dialogue}
\label{sec:method}

TalkMatrix instantiates Equation~\ref{eq:general-selection} with prompts indexed by character--situation pairs. It has two stages. First, an LLM generates multiple candidate lines independently for every pair. Second, TalkMatrix jointly selects one candidate per pair so that the complete matrix balances consistency and diversity. The second stage is the focus of our method.

Let $\mathcal{C} = \{c_1, c_2, \ldots, c_m\}$ be the set of $m$ characters and $\mathcal{S} = \{s_1, s_2, \ldots, s_n\}$ be the set of $n$ situations. We generate a dialogue matrix $X \in \mathcal{X}^{m \times n}$ where each element $x_{i,j}$ corresponds to the dialogue for the $i$-th character in the $j$-th situation. The rows of this matrix represent characters, the columns represent situations, and the elements are the dialogue themselves.
Formally,
\begin{equation}
X = \begin{bmatrix}
x_{1,1} & x_{1,2} & \cdots & x_{1,n} \\
x_{2,1} & x_{2,2} & \cdots & x_{2,n} \\
\vdots & \vdots & \ddots & \vdots \\
x_{m,1} & x_{m,2} & \cdots & x_{m,n}
\end{bmatrix},
\end{equation}
where $x_{i,j} \in \mathcal{X}$ is the dialogue for character $c_i$ in situation $s_j$.
$x_{i,j}$ is selected from a set of $k$ candidate texts $\mathcal{X}_{ij}=\{x_{i,j}^1, x_{i,j}^2, \ldots, x_{i,j}^{k}\}$ generated by an LLM. The feasible set therefore contains $k^{mn}$ matrices. We seek a high-scoring matrix under the objective below; because this search is combinatorial, our solver is not guaranteed to find its global maximum.
In the general notation of Equation~\ref{eq:general-selection}, $N=mn$, each $\mathcal{Y}_q$ is one cell-level pool $\mathcal{X}_{ij}$, and $G$ is the matrix-level objective $F$ defined in Equation~\ref{eq:obj}. The row and column interactions in $F$ make the selections non-decomposable.

\subsection{Evaluating Dialogue with Embeddings}

We represent both what a dialogue says and how it says it. Let $\phi_c$ be a content embedding and $\phi_s$ a style embedding trained to place texts with similar writing styles nearby regardless of content~\cite{qiu2025mstyledistance}. Formally,
\begin{itemize}
    \item $\phi_c: \mathcal{X} \rightarrow \mathbb{R}^{d_c}$ represents semantic content;
    \item $\phi_s: \mathcal{X} \rightarrow \mathbb{R}^{d_s}$ represents linguistic style; and
    \item $\phi(x)=[\phi_c(x);\phi_s(x)]$ is their concatenation.
\end{itemize}
We use cosine similarity $\operatorname{sim}(u,v)=\cos(u,v)$ and cosine distance $d(u,v)=1-\cos(u,v)$. Character $i$ has a reference embedding $r_c^i$, and situation $j$ has a content reference embedding $r_s^j$. The experiment section specifies the embedding models and construction of these references.

\subsection{Objectives}

We use compact notation in which the first subscript identifies the axis or reference being evaluated ($S$: situation; $C$: character), and the second identifies diversity ($D$) or consistency ($C$). Table~\ref{tab:objectives} summarizes the four requirements.
\begin{table}[t]
\centering
\adjustbox{max width=\columnwidth}{
\begin{tabular}{lll}
\toprule
Term & Requirement & Comparison \\
\midrule
$S_D$ & situation diversity & situations within a character \\
$C_D$ & character diversity & characters within a situation \\
$C_C$ & character consistency & dialogue vs. character profile \\
$S_C$ & situation consistency & dialogue vs. situation description \\
\bottomrule
\end{tabular}
}
\caption{The four matrix-level requirements. All terms are defined so that larger is better.}
\label{tab:objectives}
\end{table}

\paragraph{Situation diversity ($S_D$).}
For every character, we compute the mean pairwise content distance between that character's dialogues in different situations, then retain the worst-performing character:
\begin{equation}
S_D(X)=\min_i\frac{1}{\binom{n}{2}}\sum_{1\leq j<\ell\leq n}d\!\left(\phi_c(x_{i,j}),\phi_c(x_{i,\ell})\right).
\label{eq:sd}
\end{equation}

\paragraph{Character diversity ($C_D$).}
For every situation, we compute the mean pairwise distance between different characters and retain the least diverse situation:
\begin{equation}
C_D(X)=\min_j\frac{1}{\binom{m}{2}}\sum_{1\leq i<h\leq m}d\!\left(\phi(x_{i,j}),\phi(x_{h,j})\right).
\label{eq:cd}
\end{equation}

\paragraph{Character consistency ($C_C$).}
We compare each selected dialogue with its character reference and retain the least consistent character:
\begin{equation}
C_C(X)=\min_i\frac{1}{n}\sum_{j=1}^{n}\operatorname{sim}\!\left(\phi(x_{i,j}),r_c^i\right).
\label{eq:cc}
\end{equation}

\paragraph{Situation consistency ($S_C$).}
We compare each dialogue's content with the situation reference and retain the least consistent situation:
\begin{equation}
S_C(X)=\min_j\frac{1}{m}\sum_{i=1}^{m}\operatorname{sim}\!\left(\phi_c(x_{i,j}),r_s^j\right).
\label{eq:sc}
\end{equation}

The inner minima in Equations~\ref{eq:sd}--\ref{eq:sc} prevent a single character or situation from collapsing while the average remains high. We then combine the four terms with a second minimum so that no quality dimension can compensate for a weak one:
\begin{equation}
\begin{split}
F(X)={}&\min\!\left(w_{SD}S_D(X),w_{CD}C_D(X),\right.\\
&\left.\hspace{18mm}w_{CC}C_C(X),w_{SC}S_C(X)\right)+\lambda R(X),
\end{split}
\label{eq:obj}
\end{equation}
where the $w$ terms set the relative scales and $R$ is a local representativeness prior. For selected candidate $x_{i,j}=x_{i,j}^{a_{ij}}$,
\begin{equation}
R(X)=\frac{1}{mn}\sum_{i=1}^{m}\sum_{j=1}^{n}\frac{1}{k-1}\sum_{\ell\neq a_{ij}}\operatorname{sim}\!\left(\phi(x_{i,j}),\phi(x_{i,j}^{\ell})\right).
\label{eq:mbr}
\end{equation}
This MBR-inspired term~\cite{goel2000minimum,eikema-aziz-2020-map} discourages the selector from exploiting outlier generations that score well on a global term but are unrepresentative of their cell. We refer to the use of minima both within requirements and across requirements as the \emph{minimax/minimax} formulation. Section~\ref{sec:exp-objective-ablation} compares it with mean and harmonic-mean alternatives.

\subsection{Optimization}
\label{sec:optimization}

Exhaustive search is infeasible at the experimental scale: a $6\times6$ matrix with $k=16$ candidates per cell has $16^{36}$ possible assignments. TalkMatrix therefore treats search as an approximate inference problem over the full matrix. Our default solver is randomized multi-start coordinate ascent~\cite{Bertsekas1997}, because it is simple, exploits the matrix structure, and works well empirically. Algorithm~\ref{alg:coordinate-ascent} initializes each restart by sampling one candidate per cell. During a sweep, it visits every cell in a fixed order and replaces its current selection with the candidate that gives the highest full objective while all other cells are held fixed. After a fixed number of sweeps, it returns the highest-scoring assignment across restarts.

\begin{algorithm}[t]
\caption{Multi-start coordinate ascent for TalkMatrix}
\label{alg:coordinate-ascent}
\begin{algorithmic}[1]
\REQUIRE Candidate sets $\{\mathcal{X}_{ij}\}$, objective $F$, restarts $B$, sweeps $T$
\FOR{$b=1,\ldots,B$}
    \STATE Sample $X^{(b)}$ by choosing uniformly from every $\mathcal{X}_{ij}$
\ENDFOR
\FOR{$t=1,\ldots,T$}
    \FOR{each cell $(i,j)$ in fixed order}
        \FOR{$b=1,\ldots,B$}
            \STATE $x_{i,j}^{(b)}\leftarrow\argmax_{x\in\mathcal{X}_{ij}}F(X^{(b)}_{-ij},x)$
        \ENDFOR
    \ENDFOR
\ENDFOR
\RETURN $\argmax_{X\in\{X^{(1)},\ldots,X^{(B)}\}}F(X)$
\end{algorithmic}
\end{algorithm}

Each coordinate update is non-decreasing because the current candidate is included in the argmax. Nevertheless, coordinate ascent can terminate at a coordinate-wise local optimum, and the fixed-sweep implementation may stop before reaching one; neither variant guarantees the global maximum. Random initialization makes the algorithm stochastic. Fixing the pseudorandom seed fixes all initial assignments, and the fixed cell order and deterministic tie breaking then make an experimental run reproducible conditional on that seed.

Changing one cell affects only its character row, situation column, and local $R$ term. Our JAX implementation~\cite{jax2018github} batches all $k$ candidate evaluations and vectorizes the $B$ restarts with \texttt{vmap}, so all restart chains advance in lockstep on one GPU. This parallelization changes throughput, not the coordinate-ascent search rule.

Coordinate ascent is not intended as the uniquely best optimizer for this problem. The central modeling claim is that related prompts should be selected jointly when the collection-level objective is non-decomposable. In our instantiation, this means optimizing a matrix-level objective over multiple cells. Accordingly, Section~\ref{sec:exp-search-baselines} compares coordinate ascent with independent cell-level selection, partial row/column greedy variants, simulated annealing, and Tree-structured Parzen Estimator (TPE) sampling \cite{bergstra2011algorithms} via Optuna \cite{akiba2019optuna}.

\section{Experiment}
\label{sec:experiment}

\paragraph{Datasets.}
We construct two datasets for evaluation. 
First, we curate a set of $25$ scenarios, {\bf board game subset}, motivated from a real world game application. The game is a board game but with role-playing elements, where the non-player characters (NPCs) are designed to have distinct personalities and backgrounds, and the characters talk to the user so that the user can experience an immersive gameplay with a charming character.
For example, a user can choose to play against Zhang Fei, a historical figure from the Three Kingdoms period, who is known for his bravery. He may taunt you at the start of the game, and if you make a good move, he may praise you. If you choose to play against Zhuge Liang Kongming, a strategist from the same period. He may give you advice on how to play the game but extraordinally strong at the same time to make you feel like you are playing against a genius.
The applicational goal of the study is to generate dialogues for these characters so that users can enjoy the game with a more immersive experience.

Each scenario consists of $m=6$ characters and $n=6$ situations, with each situation corresponding to a canonical game phase (game start, deliberation, finding a good move, victory, defeat, and falling behind). The scenarios span five thematic worlds (Japanese folklore, the Warring States period, the late Edo period, Greek mythology, and the Three Kingdoms) crossed with five board and card games (chess, go, mahjong, poker, and shogi).

In addition to the board game subset, we additionally generate a {\bf synthetic subset} which consists of $50$ scenarios synthesized by GPT-4o~\cite{openai2024gpt4} designed for role-playing games in general that have a wider range of characters and situations. Each scenario consists of $5$ characters and $5$ situations.

\paragraph{Implementation of TalkMatrix.}
% \label{sec:implementation}
For each character-situation cell we generate $k=16$ candidates with \texttt{Qwen3.6-27B-FP8}\footnote{\url{https://huggingface.co/Qwen/Qwen3.6-27B-FP8}}~\cite{qwen36_27b} at temperature $1.0$. All texts are in Japanese. Character profiles and reference dialogues are themselves generated from a short seed specification. The dialogue-generation prompt is given in Appendix~\ref{apd:generation-prompt}.

For content embeddings ($\phi_c$), we use \texttt{Qwen3-Embedding-8B}\footnote{\url{https://huggingface.co/Qwen/Qwen3-Embedding-8B}} \cite{reimers-gurevych-2019-sentence}, a $4096$-dimensional multilingual text embedding model, to capture the semantic meaning of the Japanese dialogue.
For style embeddings ($\phi_s$), we use \texttt{mstyledistance},\footnote{\url{https://huggingface.co/StyleDistance/mstyledistance}} a $768$-dimensional multilingual style embedding model that embeds texts with similar writing styles closely and different styles far apart, regardless of content \cite{qiu2025mstyledistance}. 

% Unless otherwise noted, we optimize Eq.~\ref{eq:obj} with the GPU coordinate-ascent solver of Section~\ref{sec:optimization} using weights $(w_{SD},w_{CD},w_{CC},w_{SC})=(1,2,2,1)$, which emphasize character diversity and character consistency. 
We use $30$ random restarts and $8$ sweeps per restart. We fix the random seed, and hence the initial assignments, for reproducibility. 
We also show the result of Optuna TPE sampler as a domain-independent black-box optimizer as an ablation study~\cite{bergstra2011algorithms,akiba2019optuna}.

For every scenario we sample $5$ random matrices with each cell generated by \texttt{Qwen3.6-27B-FP8} at temperature $1.0$. Five randomly sampled matrices are generated for each scenario to reduce the variance. Reported random scores are averaged over these $5$ matrices.
% , each selecting one candidate per cell uniformly at random from the same candidate pool. Reported random scores are averaged over these $5$ matrices.

\paragraph{Evaluation.}
% We evaluate generated matrices both intrinsically, by the value of the objective in Eq.~\ref{eq:obj}, and extrinsically, with 
We evaluate generated matrices with an LLM-as-a-judge using \texttt{GLM-4.7-AWQ}\footnote{\url{https://huggingface.co/QuantTrio/GLM-4.7-AWQ}}~\cite{5team2025glm45agenticreasoningcoding} at temperature $0$ to score a full matrix on a $0$-$10$ scale along four rubric aspects: character-speech consistency, inter-character diversity, situation consistency, and inter-situation diversity. Then, the judge provides an overall score. 
We report the overall score as the primary metric that tests whether the optimized matrices are judged better than the random baseline. The four rubric scores are present to understand which quality dimensions are most important for judged dialogue quality. 
The judge is shown the complete $m\times n$ matrix together with a professional reference matrix, which anchors a score of $5$. The reference-anchored judge prompt is given in Appendix~\ref{apd:eval}.

\begin{table}[t]
\centering
\adjustbox{max width=\columnwidth}{
\begin{tabular}{lccc}
\toprule
Method & Objective & $\Delta$ vs.\ random & Wins \\
\midrule
Random & $0.935$ & --- & --- \\
TalkMatrix & $\mathbf{0.967}$ & $+0.031$ & $75/75$ \\
\bottomrule
\end{tabular}
}
\caption{Objective value attained by TalkMatrix (coordinate ascent, minimax/minimax) and by random selection over $75$ scenarios. Coordinate ascent obtains a higher value in every scenario; this comparison does not establish proximity to the global optimum.}
\label{tab:objective}
\end{table}

\subsection{Does Coordinate Ascent Improve the Objective over Random Selection?}
\label{sec:exp-objective}

We first test whether coordinate ascent improves Eq.~\ref{eq:obj} relative to uninformed selection. We compare the objective value of the solution found by coordinate ascent with a random-selection baseline and the Optuna TPE sampler. This experiment measures improvement over the baselines; it does not determine the global optimum or the optimality gap.

Running coordinate ascent with the minimax combination and minimax aggregation on all $75$ large-scale scenarios, the optimized matrices attain a mean objective value of $0.967 \pm 0.008$, compared with $0.935$ for the random baseline (Table~\ref{tab:objective}). The improvement is positive in \emph{every} scenario ($75/75$ wins; mean delta $+0.031$, $95\%$ CI $[+0.030, +0.032]$; one-sample $t$-test $p<10^{-80}$). Optimization takes on average $5.2$~s per scenario on a single GPU. 
Within the evaluated trial budget, the Optuna baseline does not reach the objective values found by coordinate ascent. This result shows an empirical advantage under our computational setting, but does not by itself establish that either method is near the global optimum.

\begin{table}[t]
\centering
\adjustbox{max width=\columnwidth}{
\begin{tabular}{lccc}
\toprule
 & \multicolumn{3}{c}{Aggregation} \\
\cmidrule(lr){2-4}
Combination & mean & harmonic & minimax \\
\midrule
harmonic & $-0.06$ & $+0.02$ & $+0.42$ \\
minimax  & $-0.14$ & $-0.34$ & $\mathbf{+0.98}$ \\
\bottomrule
\end{tabular}
}
\caption{Improvement in the GLM judge's overall score over the random baseline (mean over $25$ scenarios), for each combination/aggregation operator. The minimax/minimax configuration is significantly positive ($+0.98$, $17/3/5$ win/tie/loss, $p<0.002$); the others are near zero.}
\label{tab:judge-grid}
\end{table}

\subsection{Ablation Study on Multi-Objective Formulation}
\label{sec:exp-objective-ablation}
\label{sec:exp-judge}

We next evaluate the optimized matrices with the GLM judge on the $25$ curated scenarios, sweeping the combination operator $\mathrm{comb}\in\{\text{harmonic},\text{minimax}\}$ against the aggregation operator $\mathrm{agg}\in\{\text{mean},\text{harmonic},\text{minimax}\}$. Table~\ref{tab:judge-grid} reports, for each cell, the mean improvement of the optimized matrix over the random baseline in the judge's overall score.
The choice of operators matters substantially. The naive operators (means) leave the optimized matrices statistically indistinguishable from random selection. The \textbf{minimax/minimax} configuration, which maximizes the weakest objective and is most robust to a single collapsing quality dimension, performs best by a clear margin: it improves the overall score by $+0.98$ on average ($6.44$ vs.\ $5.46$; $95\%$ CI $[+0.40, +1.55]$; paired $t(24)=3.49$, $p<0.002$), winning in $17$ of $25$ scenarios (with $3$ ties). This is the objective configuration we adopt as TalkMatrix.
\begin{table}[t]
\centering
\adjustbox{max width=\columnwidth}{
\begin{tabular}{lccc}
\toprule
Method & Overall & $\Delta$ vs.\ random & W/T/L \\
\midrule
Cell situation & 4.84 & $-0.62$ & 9/1/15 \\
Cell character & 5.48 & $+0.02$ & 12/0/13 \\
Cell char.+sit.+MBR & 5.48 & $+0.02$ & 9/2/14 \\
Cell MBR & 6.04 & $+0.58$ & 15/3/7 \\
Row greedy & 5.60 & $+0.14$ & 12/0/13 \\
Simulated annealing & 6.20 & $+0.74$ & 18/2/5 \\
Column greedy & 6.32 & $+0.86$ & 19/1/5 \\
TalkMatrix & \textbf{6.44} & $\mathbf{+0.98}$ & 17/3/5 \\
\bottomrule
\end{tabular}
}
\caption{Selection-policy and search baselines on the $25$ curated scenarios, evaluated by the GLM judge. Random selection has mean overall score $5.46$. Cell methods select each candidate independently; row/column greedy optimize one row or column at a time; TalkMatrix uses full-matrix coordinate ascent.}
\label{tab:selection-baselines}
\end{table}

\subsection{Selection-Policy and Search Baselines}
\label{sec:exp-search-baselines}

We next test whether the benefit comes from global matrix-level selection or from simpler local reranking. Table~\ref{tab:selection-baselines} compares TalkMatrix with independent cell-level methods, partial-matrix greedy variants, and simulated annealing on the same $25$ curated scenarios and GLM-judge protocol. The cell-level baselines choose each candidate independently using situation consistency, character consistency, their combination with MBR, or MBR alone. The row- and column-greedy baselines optimize multiple cells at a time but do so incrementally: row greedy constructs one character row at a time, whereas column greedy constructs one situation column at a time. Simulated annealing is a generic combinatorial optimizer using random single-cell mutations.

The results support two conclusions. First, independent consistency-based reranking is not sufficient: cell situation selection falls below random ($\Delta=-0.62$), and cell character or cell character+situation+MBR selection is nearly identical to random ($\Delta=+0.02$). Cell MBR is stronger ($\Delta=+0.58$), suggesting that avoiding unrepresentative outlier generations is useful even without global coordination. Second, optimizing multiple cells under the matrix-level objective is important. Simulated annealing improves substantially over random ($\Delta=+0.74$), and column greedy is close to full TalkMatrix ($+0.86$ vs.\ $+0.98$). Thus coordinate ascent is one effective search procedure, but the broader lesson is that the objective should be optimized over a coupled matrix rather than independently per prompt. The strong column-greedy result also suggests a practical online variant for settings where dialogue for a new situation or game phase is added incrementally.

\subsection{Comparison with a Black-Box Optimizer}
We also compare coordinate ascent with the Optuna TPE sampler~\cite{bergstra2011algorithms,akiba2019optuna}, a widely used black-box optimization method. On $50$ synthetic scenarios with $1{,}000$ trials per scenario, Optuna takes a median of $586.6$~seconds per scenario and approximately $16.3$~hours in total, compared with $5.2$~s per scenario for coordinate ascent. Note that the walltime is subject to the computational resources available. Optuna involves many CPU operations whereas coordinate ascent is fully GPU-parallelized, so the wall-clock time is not directly comparable. With the minimax objective, Optuna-selected matrices are statistically indistinguishable from random ($6.50$ vs.\ $6.46$; mean difference $+0.04$; $95\%$ CI $[-0.199,+0.279]$; $p=0.7407$; $32/37/31$ wins/ties/losses). With the weighted-sum objective, Optuna-selected matrices score lower than random selection under the GLM judge ($6.13$ vs.\ $6.46$; mean difference $-0.33$; $95\%$ CI $[-0.608,-0.052]$; paired $t$-test $p=0.0204$; $26/25/49$ wins/ties/losses). These results indicate that generic black-box sampling is not enough under this budget, while structure-aware or matrix-level search methods can be effective.

% Together with the intrinsic comparison in Section~\ref{sec:exp-objective}, these results indicate that black-box TPE sampling and a naive weighted sum do not account for TalkMatrix's improvement.

\subsection{Which Requirements Matter? Objective-Component Ablation}
\label{sec:exp-requirement-ablation}

After fixing the objective-composition operator to the best-performing setting (minimax combination with minimax aggregation), we analyze which requirement components are most important for dialogue quality. Unlike operator ablations, which probe optimization design choices, this study asks a substantive NLP question: which real-world constraints should be prioritized in role-playing dialogue generation.

We evaluate nine methods on the same $25$ curated scenarios and with the same GLM-judge protocol as Section~\ref{sec:exp-judge}: the full objective, four drop-one variants (drop SD, drop SC, drop CD, drop CC), and four single-term variants (single SD, single SC, single CD, single CC). Table~\ref{tab:requirement-ablation} reports the mean overall score, mean gain over random, scenario-level wins/ties/losses against random, and trial-level win rate with ties counted as $0.5$.

The results highlight three patterns. First, CD is the most critical component: dropping CD is the only setting that falls below random on average ($\Delta=-0.10$) and shows the largest gap to the full objective ($-1.08$). Second, SC is the strongest single component: single SC is closest to full both in mean gain ($-0.24$ gap) and tie-adjusted trial win rate ($0.676$ vs. $0.688$). Third, SD and CC are beneficial but less decisive than CD and SC: removing SD or CC degrades performance, yet both remain above random.

\begin{table}[t]
\centering
\adjustbox{max width=\columnwidth}{
\small
\begin{tabular}{lccccc}
\toprule
Method & Overall & $\Delta$ vs rand & $\Delta$ vs full & W/T/L & Win-rate$_{0.5}$ \\
\midrule
Full      & 6.44 & +0.98 & 0.00  & 17/3/5  & 0.688 \\
Drop SD   & 6.12 & +0.66 & -0.32 & 16/2/7  & 0.648 \\
Drop SC   & 5.76 & +0.30 & -0.68 & 13/3/9  & 0.556 \\
Drop CD   & 5.36 & -0.10 & -1.08 & 10/5/10 & 0.496 \\
Drop CC   & 6.12 & +0.66 & -0.32 & 16/2/7  & 0.632 \\
Single SD & 5.92 & +0.46 & -0.52 & 13/5/7  & 0.584 \\
Single SC & 6.20 & +0.74 & -0.24 & 16/2/7  & 0.676 \\
Single CD & 6.04 & +0.58 & -0.40 & 15/3/7  & 0.624 \\
Single CC & 5.84 & +0.38 & -0.60 & 13/3/9  & 0.560 \\
\bottomrule
\end{tabular}
}
\caption{Objective-component ablation on $25$ scenarios (GLM judge). Overall is mean judge score. Win-rate$_{0.5}$ is trial-level win rate against random with ties counted as $0.5$.}
\label{tab:requirement-ablation}
\end{table}

Taken together, these findings suggest that requirement selection is a central modeling issue rather than only an optimization detail. In particular, preserving CD and SC appears most important for judged dialogue quality in this role-playing setting.

\subsection{Lexical Diversity}
\label{sec:lexical-diversity}

We additionally evaluate surface-level diversity using lexical overlap metrics as a supplementary evaluation. On each of the $25$ curated scenarios, we compare the TalkMatrix output with the same five random matrices used in the main evaluation. For each character, we tokenize its six Japanese dialogues using MeCab through Fugashi with UniDic Lite and remove punctuation~\cite{mccann-2020-fugashi}. We compute corpus-level Distinct-1 and Distinct-2~\cite{li-etal-2016-diversity} as the proportion of unique unigrams or bigrams among all within-dialogue $n$-grams for that character; $n$-grams do not cross dialogue boundaries. Character scores are averaged within each scenario, and the random score is averaged over the five random matrices.

\begin{table}[t]
\centering
\adjustbox{max width=\columnwidth}{
\begin{tabular}{lcccc}
\toprule
Metric & TalkMatrix & Random & Improvement & 95\% CI \\
\midrule
Distinct-1 $\uparrow$   & 0.6370 & 0.6248 & +0.0122 & [0.0057, 0.0182] \\
Distinct-2 $\uparrow$   & 0.9696 & 0.9640 & +0.0056 & [0.0026, 0.0084] \\
\bottomrule
\end{tabular}
}
\caption{Lexical diversity analysis on the $25$ curated scenarios. Confidence intervals are scenario-bootstrap intervals.}
\label{tab:lexical-diversity}
\end{table}

Table~\ref{tab:lexical-diversity} shows the score. TalkMatrix improves Distinct-1 and Distinct-2, with positive differences in $21/25$ and $20/25$ scenarios, respectively.
The result indicates that TalkMatrix improves lexical diversity and present users more varied dialogues than random selection.

\section{Conclusions}
We formulated structured multi-prompt, multi-completion selection, in which one completion is chosen from each prompt's candidate pool under a non-decomposable collection-level objective. TalkMatrix instantiates this problem for character dialogue by jointly selecting lines across a character--situation matrix. It combines four matrix-level consistency and diversity requirements with a two-level minimax objective and approximately optimizes the result with GPU-parallel, multi-start coordinate ascent. Across $75$ scenarios, coordinate ascent attains higher objective values than random selection, and on $25$ curated role-playing scenarios the selected matrices are judged significantly better than the random baseline by a language-model judge. Crucially, judged improvement depends on both objective design and search scope: the minimax formulation protects the weakest dimension, and methods that optimize the coupled matrix outperform independent cell-level consistency reranking. Coordinate ascent is an effective default solver, while simulated annealing and column-wise greedy search show that other matrix-level or partial-matrix search strategies can also be useful. These experiments validate the general formulation only in character dialogue; applying structured selection to other multi-output tasks remains future work.

Future work will tune the objective weights and operators against human judgments so that the trade-off between the objectives is set optimally with respect to human preferences, and will extend the approach to incorporate the objectives directly into the generation stage rather than as a post-hoc selection step.

% \bibliography{anthology-1,anthology-2,ms,ms2,story}

\input{AnonymousSubmission2027.bbl}
% \clearpage
\appendix

\input{prompts_appendix.tex}

% \section{Generation Examples}
% \label{apd:generation_examples}
\input{sangokushi_go_example.tex}

% Check whether the conference requires a reproducibility checklist to be included in the paper.
% If so, you can uncomment the following line and ajust the path to include it.
% \input{ReproducibilityChecklist.tex}

\end{document}

%% file: prompts_appendix.tex
\section{Prompts}
\label{apd:prompts}

The prompts we use in the experiment are in Japanese because all dialogue is Japanese.
For accessibility, we provide English translations by GPT-5.5 below.
Braced expressions denote values inserted by the implementation.

\subsection{Dialogue Candidate Generation Prompt}
\label{apd:dataset}
\label{apd:generation-prompt}

For each character--situation cell, the model is sampled repeatedly with the following messages to obtain the candidate pool.

\paragraph{System message.}
\begin{examplebox}
\begin{Verbatim}[fontsize=\scriptsize,breaklines=true,breakanywhere=true]
You are a scriptwriter for stories in role-playing games.
Generate a dialogue line based on the following character specification.

## Specification of {character_name}
{character_description}

## Output format
Role-play the character and output only the character's utterance.
\end{Verbatim}
\end{examplebox}

\paragraph{User message.}
\begin{examplebox}
\begin{Verbatim}[fontsize=\scriptsize,breaklines=true,breakanywhere=true]
Role-play {character_name} in the scene "{situation_name}" and
generate one dialogue line.

## Specification of {situation_name}
{situation_description}
\end{Verbatim}
\end{examplebox}

\subsection{Full-Matrix Judge Prompt}
\label{apd:eval}

The judge receives a professional reference matrix in the system message and the matrix under evaluation in the user message. The reference matrix is defined as score 5, and all scores are assigned relative to it. The judge temperature is zero.

\paragraph{System message.}
\begin{examplebox}
\begin{Verbatim}[fontsize=\scriptsize,breaklines=true,breakanywhere=true]
You are an expert in evaluating stories and character specifications.
Strictly evaluate the provided dialogue matrix from a fair and objective
perspective.

## Evaluation task

Evaluate, on a 1--10 scale, how well the evaluation matrix expresses the
character and situation specifications.

## Evaluation criteria

Evaluate the matrix from the following four perspectives.

1. Character speech consistency: Are the distinctive features of each
   character's speech consistent across all situations?
2. Inter-character diversity: Are the characters clearly differentiated in
   speaking style and values, with distinctive personalities?
3. Situation consistency: Is each dialogue line a natural and appropriate
   response to its situation?
4. Inter-situation diversity: Does each character's dialogue change
   appropriately across different situations?

## Evaluation scale

Assign scores from 0 to 10.

- 10: Exceeds a product level quality on every criterion.
- 8: Close to a product level quality on every criterion.
- 5: High quality on every criterion, but not at a product level.
- 2: Low quality in regard to one of the criteria.
- 0: The matrix has a major flaw.

## Evaluation procedure

1. Analyze the strengths and areas for improvement according to the
   evaluation criteria above. This analysis provides the basis for the
   scores.
2. Assign an individual score from 0 to 10 for each criterion.
3. Based on the analysis, jointly consider the four criteria and assign a
   final overall score.

## Output format

Analysis:
[Write your analysis according to the evaluation criteria.]

Scores by criterion:
1. Character speech consistency: [[integer score from 0 to 10 only]]
2. Inter-character diversity: [[integer score from 0 to 10 only]]
3. Situation consistency: [[integer score from 0 to 10 only]]
4. Inter-situation diversity: [[integer score from 0 to 10 only]]

Overall score:
[[integer score from 0 to 10 only]]
\end{Verbatim}
\end{examplebox}

% \paragraph{User message.}

% The matrix serializer repeats the following block for every situation and character, ensuring that the complete matrix is shown to the judge:

% \begin{Verbatim}[fontsize=\scriptsize,breaklines=true,breakanywhere=true]
% # Situation: {situation_name}

% ## Situation description:
% {situation_description}

% ## {character_name}:
% {selected_dialogue}
% \end{Verbatim}

%% file: sangokushi_go_example.tex
% \subsection{Complete Example: Heroes of the Three Kingdoms Playing Go}
% \label{apd:sangokushi-go}

This section shows one complete $6\times6$ instance, translated from Japanese into English for readability using GPT-5.5. The setting combines six figures from the Chinese Three Kingdoms period with six moments in a game of Go. We compare TalkMatrix (coordinate ascent, minimax/minimax) with one uniformly sampled baseline (random matrix 0). The translations preserve the meaning, register, and occasional defects of the Japanese rather than correcting them after selection.

\paragraph{Character specifications.}
\begin{table*}[t]
\centering\scriptsize
\begin{examplebox}
\begin{tabularx}{\linewidth}{@{}lX@{}}
\toprule Character & English translation of the specification \\
\midrule
Liu Bei & A descendant of the Han imperial clan who founded Shu Han. Benevolent and sympathetic to the people, he prizes loyalty and family bonds and has the resolve and charisma to unite others in adversity. He speaks sincerely and gravely, respecting others while stating his convictions quietly but firmly. \\
Guan Yu & A renowned Shu Han general and sworn brother of Liu Bei and Zhang Fei, later deified as Lord Guan. Loyal, frugal, honorable, and proud, he prizes ritual and fidelity. His archaic, dignified speech is confident and sometimes haughty. \\
Zhang Fei & A renowned Shu Han general and sworn brother of Liu Bei and Guan Yu. Bold, fierce, quick-tempered, and terrifying in battle, yet loyal and generous to his men, he speaks directly and forcefully and becomes intimidating when excited. \\
Zhuge Liang & The brilliant, cautious, diligent, and incorruptible chancellor of Shu Han, known as the ``Crouching Dragon.'' Deeply loyal to Liu Bei, he speaks calmly and learnedly, with respect and firm conviction. \\
Cao Cao & A powerful late-Han statesman and founder of Wei. Ambitious and pragmatic, receptive to talent but deeply suspicious, he can make ruthless decisions for the larger objective. He speaks with concise authority, often using commands and rhetorical questions. \\
Sun Quan & Successor to the Sun family in Jiangdong and later emperor of Wu. Calm and perceptive, he excels at appointing talent and balancing power. A flexible pragmatist, he speaks in a dignified, deliberate, and composed archaic register. \\
\bottomrule
\end{tabularx}
\end{examplebox}
\caption{Character specifications for the Three Kingdoms--Go example, translated from the generation input.}
\label{tab:sangokushi-characters}
\end{table*}

\paragraph{Situation specifications.}
\begin{examplebox}
The six situations are: \textbf{Start of match}, when the players exchange greetings and place the first stone; \textbf{Thinking}, when a player surveys the board and considers the next move; \textbf{Finding a good move}, the instant a promising move occurs to the player; \textbf{Victory}, the instant the game is won; \textbf{Defeat}, the instant the game is lost; and \textbf{Disadvantage}, the instant the player realizes that the position has become unfavorable.
\end{examplebox}

\paragraph{Selected matrices.}
Tables~\ref{tab:sango-1}--\ref{tab:sango-6} show every cell of both matrices. A dagger marks identity confusion discussed below.

\begin{table*}[t]\centering\scriptsize
\begin{examplebox}
\begin{tabularx}{\linewidth}{@{}lXX@{}}\toprule Character & TalkMatrix & Random selection \\\midrule
Liu Bei & ``I accept. In this match, on which the peace of the realm rests, I humbly ask for your lasting goodwill.'' & ``Kongming, Zhuge, it is a heartfelt honor to play you. This is no mere contest, but a fateful game for peace in the realm and relief for its people. Let us respect one another, calm our hearts, and make the first move.''$^\dagger$ \\
Guan Yu & ``I accept. With this single black stone, I now contest the realm with you.'' & ``We exchange courtesies. Clouds and mist rise over the board, and a once-in-a-lifetime chance approaches. I shall receive your move with my mind's eye. Let us reveal our full mastery to the board's edge. Now, the first move.'' \\
Zhang Fei & ``Bwahaha! To blazes with courtesy! This board is no different from the thrust and parry of my spear! The first move---take that, black stone!'' & ``Ha! Come at me, enemy commander! This will not end as a mere game of stones. With this move, I shall take your general's head!'' \\
Zhuge Liang & ``The courtesies are complete. Now, let us inscribe the first star upon this nineteen-line world.'' & ``Then, the first move. With the energy of Heaven, I shall scatter stars across this board.'' \\
Cao Cao & ``Hmph. Let the contest begin.'' & ``Hahaha! Even dominion over the realm is compressed into this nineteen-line grid. Do you think your white stones can stop my black? Try me!'' \\
Sun Quan & ``I come with due respect. Let me see how brightly your strategy shines upon this board.'' & ``Accepted. In place of further ceremony, the first move. From here begins the game in which We shall build Our realm.'' \\
\bottomrule\end{tabularx}
\end{examplebox}
\caption{Complete matrices for \emph{Start of match}.}\label{tab:sango-1}\end{table*}

\begin{table*}[t]\centering\scriptsize
\begin{examplebox}
\begin{tabularx}{\linewidth}{@{}lXX@{}}\toprule Character & TalkMatrix & Random selection \\\midrule
Liu Bei & ``Hmm. Following Kongming's counsel and considering the whole board, should I yield a step here? No---the time is ripe.'' & ``Hmm. The left frontier is secure, but the open space on the right troubles me. Would playing here form the position that decides the realm? No---perhaps haste makes waste.'' \\
Guan Yu & ``An unseen presence swirls in the board's depths. Where must I play to seize this current all at once?'' & ``An unseen presence swirls in the board's depths. Where must I play to seize this current all at once?'' \\
Zhang Fei & ``The enemy is closing in here. Then I'll strike the reverse! Here---I'll slam it down with a clang!'' & ``Should I close the tiger's mouth, or attach and charge into his thickness? Hmph! My instinct says one stone here will smash his forces!'' \\
Zhuge Liang & ``A gap in the enemy formation. No, wait: trap or opportunity? Wind, terrain, and the enemy's mind intertwine, yet the sole chance lies in this corner. I see: still waters run deep. This stone shall reverse the board's momentum.'' & ``The position is becoming clear, yet that move leaves an opening. Read the opponent's heart, avoid provoking him, and do not miss the chance... Good, here.'' \\
Cao Cao & ``White's thickness may soon be complete. Press on the left? No, too hasty. Looking over the whole board, the move that can inflict the greatest blow is... here.'' & ``White's thickness may soon be complete. Press on the left? No, too hasty. Looking over the whole board, the move that can inflict the greatest blow is... here.'' \\
Sun Quan & ``Wei is contained and Wu is stable. Should the next move fill this gap, or reach west to check Shu? Only one point can kill both birds with one stone.'' & ``Viewed as a whole, White's left-side thickness exerts pressure. But if I concede the right corner, I can win the center fight. Good---I shall enclose here.'' \\
\bottomrule\end{tabularx}
\end{examplebox}
\caption{Complete matrices for \emph{Thinking}.}\label{tab:sango-2}\end{table*}

\begin{table*}[t]\centering\scriptsize
\begin{examplebox}
\begin{tabularx}{\linewidth}{@{}lXX@{}}\toprule Character & TalkMatrix & Random selection \\\midrule
Liu Bei & ``Kongming, what say you to this plan? For the people of the realm, I have an idea that seems good for now.'' & ``Ah! I see! This is it!'' \\
Guan Yu & ``Hmph. Will this not make matters interesting? Let us set this stratagem and exploit the enemy commander's carelessness.'' & ``An ingenious plan occurs to me. This is the perfect scheme to catch Cao Cao in one sweep!'' \\
Zhang Fei & ``Hahaha! A perfect opening granted by Heaven! Not even strategist Kongming could misread this. Brother, move the ambush at once!'' & ``Damn you, Zhang Fei---I've got it! This plan will smash Cao Cao!''$^\dagger$ \\
Zhuge Liang & ``Now I may preserve a strategist's honor. The plan to deceive Heaven is complete.'' & ``If I make an Empty Fort in this corner and stir the enemy's suspicion, it may settle everything. May Heaven aid us---Kongming shall do it.'' \\
Cao Cao & ``This is it. I shall seal Yuan Shao's retreat completely. This move decides the first step toward unifying the realm!'' & ``Splendid! To think there was a move that struck this vulnerable point. Mengde, this move proves one fit to gaze down upon the realm!'' \\
Sun Quan & ``Let me ask that old Cao Cao one thing: can his men cross the waves of the Yangtze?'' & ``Let me ask that old Cao Cao one thing: can his men cross the waves of the Yangtze?'' \\
\bottomrule\end{tabularx}
\end{examplebox}
\caption{Complete matrices for \emph{Finding a good move}.}\label{tab:sango-3}\end{table*}

\begin{table*}[t]\centering\scriptsize
\begin{examplebox}
\begin{tabularx}{\linewidth}{@{}lXX@{}}\toprule Character & TalkMatrix & Random selection \\\midrule
Liu Bei & ``With Heaven's aid, we have broken their formation. Officers and soldiers, well done. Today we have truly shown the world the great Han's purpose.'' & ``Heaven has blessed us! Your sincerity and valor brought this victory. Let us rejoice and walk onward together.'' \\
Guan Yu & ``Even dust, piled high, becomes a mountain. Your skill was respectable, but still no match for my Green Dragon Crescent Blade. Know this as the victory of one who kept faith.'' & ``The outcome was decided from the start.'' \\
Zhang Fei & ``Hahaha! Felled like a wounded rat yet again! Before the spear point of Zhang Yide, no enemy remains!'' & ``Hahaha! Exhilarating! Enemy commanders, it is only natural that you submit before Zhang Yide of Yan!'' \\
Zhuge Liang & ``One who disrupts the game record of the realm should not sit before me. Next time, I shall shatter Wei's army itself upon this board.'' & ``The threads of fate moved exactly as we wished. Soldiers of Wei, shed your armor and kneel before the benevolence of Shu.'' \\
Cao Cao & ``Hahaha! Dominion over the realm now rests squarely in the palm of Cao Cao!'' & ``The hero of the realm is the victor of this game!'' \\
Sun Quan & ``Well, Kongming? That move was harsher than expected, was it not? Victory belongs to one who never misses a momentary opening---and I seized it.'' & ``Even the Yangtze wind now favors us. Zhou Yu and Zhang Zhao, I reward your loyalty. This victory shall make Wu's foundations unshakable.'' \\
\bottomrule\end{tabularx}
\end{examplebox}
\caption{Complete matrices for \emph{Victory}.}\label{tab:sango-4}\end{table*}

\begin{table*}[t]\centering\scriptsize
\begin{examplebox}
\begin{tabularx}{\linewidth}{@{}lXX@{}}\toprule Character & TalkMatrix & Random selection \\\midrule
Liu Bei & ``Hahaha... Does Heaven still reserve a mission for Liu Bei, or is this where I meet my end?'' & ``So this is Heaven's decree. There is nothing left to regret or mourn.'' \\
Guan Yu & ``Gah! What caught me was not your honor, but your skill...'' & ``Ah... Is this a welcome rain? It seems even Guan has again had the ground taken from beneath him.'' \\
Zhang Fei & ``Hahaha! So close! Yet is this my limit? Next time I shall surely take your head!'' & ``Damn it! Just you wait. Next time Zhang Yide's three thousand cavalry will grind even your bones to dust!'' \\
Zhuge Liang & ``I overlooked that move. Sima Zhongda, make this defeat nourishment for the years ahead.'' & ``How regrettable. Though Heaven's opportunity is not yet gone, human calculation is exhausted. Sima Zhongda, I sincerely congratulate you on your victory.'' \\
Cao Cao & ``Hahaha! So that was the move. The true law of Heaven surpasses my calculations. A pity, but exhilarating!'' & ``Is it Heaven's will, or has my fortune run out? Losing one game does not surrender the realm. Next time I shall swallow you whole!'' \\
Sun Quan & ``Unlike the flames at Red Cliffs, this is a cold defeat brought by my own hand. To Cao Aman, Liu Jingyuan, and you: by the pride of Jiangdong, I shall repay this humiliation.'' & ``Understood. This loss will not shake Wu's morale. Tomorrow we devise a new plan.'' \\
\bottomrule\end{tabularx}
\end{examplebox}
\caption{Complete matrices for \emph{Defeat}.}\label{tab:sango-5}\end{table*}

\begin{table*}[t]\centering\scriptsize
\begin{examplebox}
\begin{tabularx}{\linewidth}{@{}lXX@{}}\toprule Character & TalkMatrix & Random selection \\\midrule
Liu Bei & ``What! Such skill... Has the opening in our formation already been pierced? Kongming, what do you make of this?'' & ``Ah... So this move had already seen through Our plan.'' \\
Guan Yu & ``My Green Dragon is blocked by layer upon layer of enemy blades. Yet I must not panic; now is the time to step back and read the wind.'' & ``This shape is perilous. To be engulfed by the enemy's scheme... Yet my loyalty shall not waver! Now I must play the move that reverses it!'' \\
Zhang Fei & ``What! Who in blazes are you? My spear point cannot possibly have been turned aside!'' & ``Oh---did I let that slip? No, this is bad!'' \\
Zhuge Liang & ``Whose move is it that calls forth death? No need to ask: at this instant, the opponent has read my thoughts completely.'' & ``Not even a fragment of an opening appears in the enemy commander's play. Is our formation already collapsing? I must devise a plan at once.'' \\
Cao Cao & ``The enemy commander's gaze, the troops' formation, even the wind now seem to glare at us. Is this not mere defeat, but Heaven itself intent on stopping Cao Mengde?'' & ``You intend to trap this old body? A splendid move, but that stone will become the final stone sealing your fate.'' \\
Sun Quan & ``Your countless forces are already swallowing my board. Impressive, but the game is not over. Watch this move of reversal!'' & ``Somehow the air has grown heavy. What is this---has my hand become blurred?'' \\
\bottomrule\end{tabularx}
\end{examplebox}
\caption{Complete matrices for \emph{Disadvantage}.}\label{tab:sango-6}\end{table*}

\paragraph{A visible failure of independent random selection.}
In the random opening line, Liu Bei addresses ``Kongming'' and ``Zhuge'' as though they were two people, although Kongming is Zhuge Liang's courtesy name. The random ``good move'' line for Zhang Fei likewise addresses Zhang Fei by name, creating an awkward self-reference. These local entity errors are present in the sampled candidates and are not repaired by choosing each cell independently. Several random lines also replace the Go situation with literal battlefield narration. TalkMatrix is not an error-correction system and its selected matrix still contains martial metaphors, but joint selection avoids these two conspicuous identity confusions in this example.